\documentclass[runningheads]{llncs}
\usepackage{cite}
\usepackage[T1]{fontenc}
\usepackage{graphicx}
\usepackage{cleveref}

\begin{document}
\title{Hint of Pseudo Code (HoPC):
Zero-Shot Step by Step Pseudo Code Reasoning Prompting}
\titlerunning{Hint of Pseudo Code}
%
\author{Iok Tong Lei\inst{1}\orcidID{0009-0007-9431-2740}  \and
Ziyu Zhu \inst{1}\orcidID{0000-0003-1556-0791} \and
Han Yu \inst{1}\orcidID{0009-0000-2828-4541} \and 
Yige Yao \inst{1}\orcidID{0009-0004-2886-6874} \and
Zhidong Deng \inst{*,1}\orcidID{0000-0001-9970-1023}}

\authorrunning{Iok Tong Lei et al.}
%
\institute{\textsuperscript{1}Department of Computer Science and Technology, 
Institute for Artificial Intelligence at Tsinghua University (THUAI), 
Beijing National Research Center for Information Science and Technology (BNRist), Tsinghua University, Beijing 100084, China
\email{lixt22@mails.tsinghua.edu.cn}\\
\email{michael@tsinghua.edu.cn}\\}
\maketitle              
\begin{abstract}
    Prompting a language model (LM) is an increasingly important research topic for better utilization of large language models (LLMs). While simple prompting is effective for single-step questions, it fails to activate the correct knowledge path for multi-step reasoning tasks consistently. The few-shot Chain of Thought (CoT), serves as an advanced prompting strategy that explains and demonstrates the reasoning process to the LLM, outperforming simple prompting in challenging reasoning tasks such as arithmetic and common-sense reasoning. The Program of Thought (PoT) aims to generate text and programming language solutions for multi-step reasoning problems. In zero-shot CoT, the prompt is simply ``Let's think step by step'', which is overly simplistic and does not adequately demonstrate a robust reasoning process for complex reasoning challenges. Additionally, PoT requires an extra interpreter to execute the answer and struggles with semantic reasoning problems like StrategyQA. This paper introduces a novel Hint of Pseudo Code (HoPC) prompting technique that does not require extra interpreter as in PoT and incorporates a more powerful zero-shot problem decomposition and semantic code reasoning capabilities than zero-shot CoT. It consists of three components: problem decomposition, semantic code reasoning, and answer extraction. 
    We prompt these components as hints in a sequential, step by step manner, making it easy to tailor and explain for various tasks. Our experimental results on various LLMs indicate that HoPC prompting outperforms zero-shot CoT in zero-shot reasoning tasks. We conduct experiments on mathematical tasks, including GSM8K, ADDSUB, AQUA, SVAMP, and common-sense tasks such as StrategyQA. Notably, the accuracy of the proposed HoPC prompting implemented on Llama3-8b improved from 51.8\% to 63.5\% on GSM8K, from 37.4\% to 74.4\% on AQUA, from 51.8\% to 75.5\% on SVAMP, and from 53.6\% to 74.9\% on ADDSUB.

\keywords{Prompting \and LLMs \and Reasoning}
\end{abstract}
\section{Introduction}

  In recent years, large language models achieved great success in various reasoning tasks. With zero-shot reasoning, the proposed zero-shot Chain of Thought (CoT) \cite{NEURIPS2022_8bb0d291} demonstrates that LLMs are effective zero-shot reasoners by simply adding ``let’s think step by step''. However, the LLM generates the solving process without formatted instructions on how to perform logical reasoning. This prompt encourages the LLM to decompose the problem, but stimulates the reasoning process naively. The zero-shot Program of Thought (PoT) \cite{chen2023program} performs better by utilizing Python as an additional tool for mathematical reasoning. However, it can only focus on mathematical reasoning tasks due to lack of semantic reasoning ability. Both studies highlight the challenges of zero-shot reasoning tasks with LLMs.
 
\begin{figure}[htbp]
\centerline{\includegraphics[scale=0.4]{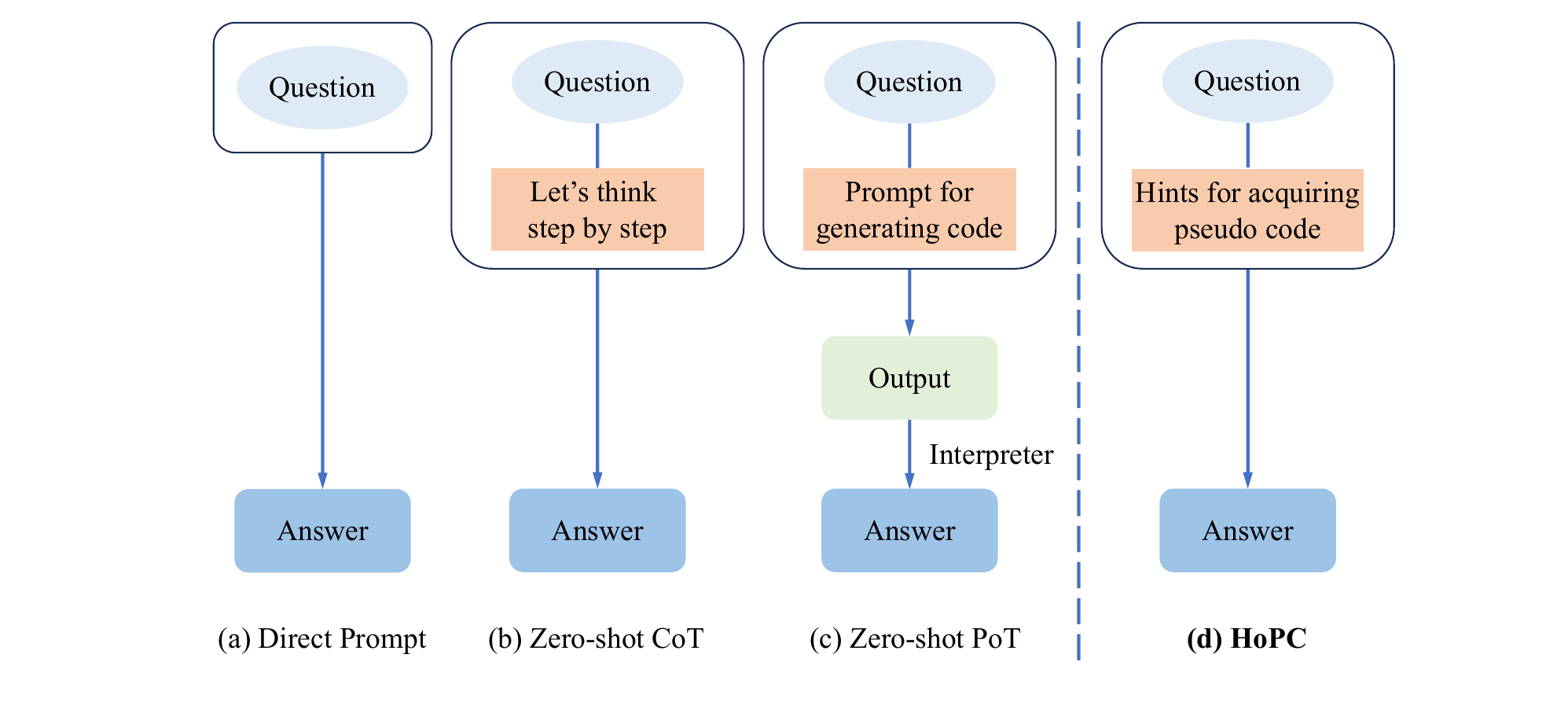}}
\caption{Different structures of zero-shot prompting. (a) Direct prompt to the LLM. (b) Zero-shot CoT prompting shows the ability of ``Let's think step by step'' \cite{NEURIPS2022_8bb0d291}. (c) Zero-shot PoT shows a ability of prompting to generating code \cite{chen2023program}. (d) Our HoPC follows the structure of zero-shot CoT but performs a better reasoning process.}
\label{fig1}
\end{figure}

\begin{figure}[htbp]
\centerline{\includegraphics[scale=0.4]{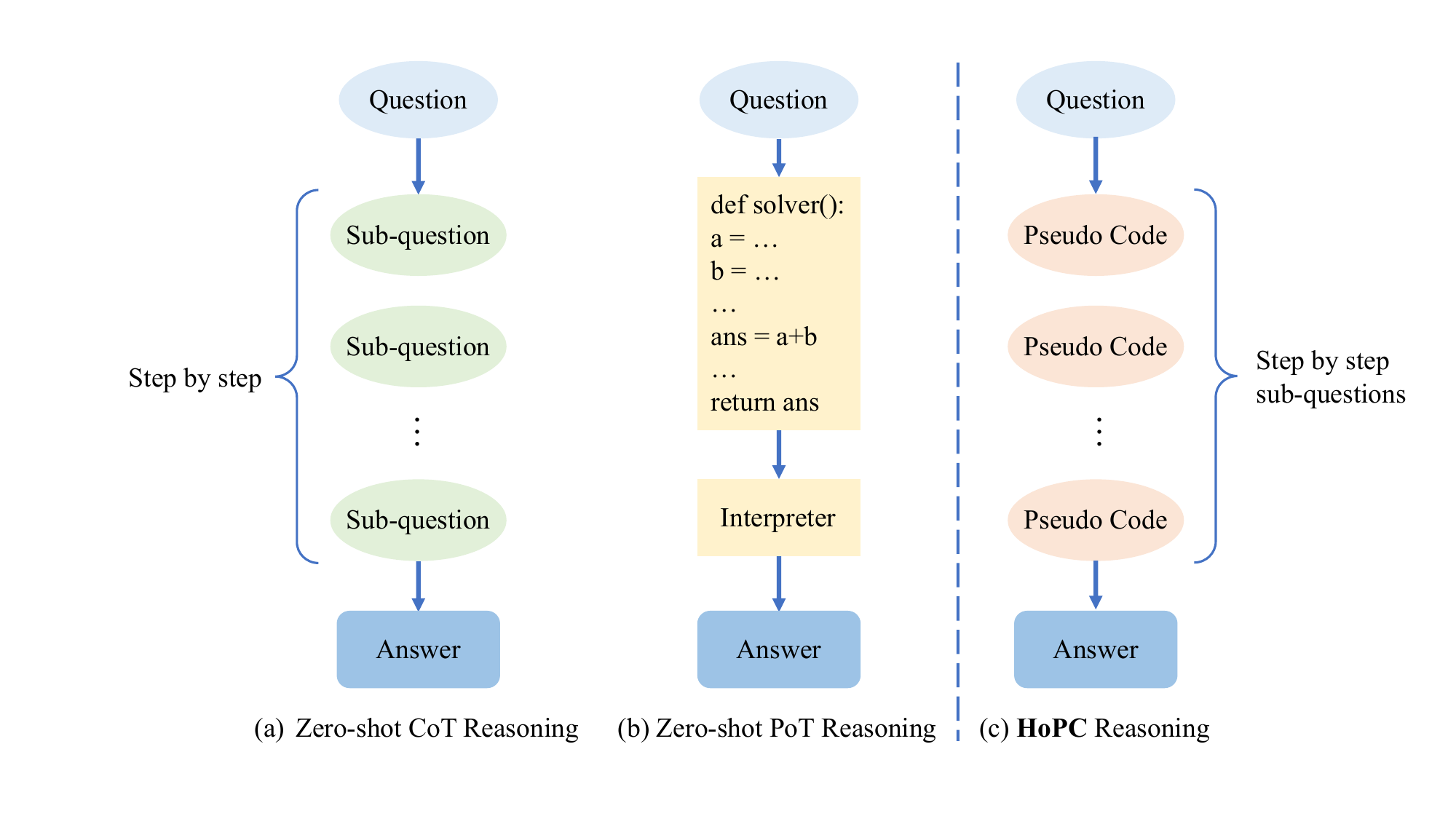}}
\caption{Different reasoning process of zero-shot prompting. (a) Zero-shot CoT does step by step problem decomposition \cite{NEURIPS2022_8bb0d291}. (b) Zero-shot PoT does code reasoning \cite{chen2023program}. (c) Our HoPC does step by step Pseudo code reasoning.}
\label{fig2}
\end{figure}

Although zero-shot PoT \cite{chen2023program} struggles in semantic reasoning, the code shows good performance in logical reasoning. Code provides a well-designed structure to implement complex algorithms for difficult problems \cite{chen2023program}. Encouraging LLMs to generate code for complex problems is more manageable when LLMs are directly encouraged to generate the reasoning process. LLMs are designed to predict the next token, but do not solve reasoning problems. However, some reasoning problems require semantic understanding, such as finding all jokes within an essay. There is no way to write a function to identify jokes with only programming languages. Therefore, to do semantic code reasoning, we propose general pseudo code that can be quickly executed through LLMs as an interpreter. 

We define general pseudo code as a high-level, language-agnostic description of a process or algorithm, designed to illustrate the logical flow of a solution without strict adherence to control structures or programming syntax.

 To build a zero-shot prompt method capable of problem decomposition and code reasoning, we propose the Hint of Pseudo Code (HoPC) prompting technique. We use the traditional Chain of Thought (CoT) as an activation path that generates the correct reasoning path from large language models (LLMs). We introduce hints, referred to as a hint chain, which are intuitive to humans and efficient for LLMs to process. This chain consists of three main parts: (1) asking for sub-questions; (2) requesting general pseudo code to tackle these sub-questions; and (3) obtaining the answer in a specific manner. These three parts automatically guide the LLM to perform zero-shot reasoning. The output from the sub-questions not only guides the LLMs' reasoning but also aids users in understanding the LLM's ``mind map''. At the same time, the general pseudo code provides a more precise and logical reasoning process to guide the LLM in generating a more logical reasoning path. The key idea of HoPC is to encourage LLMs to produce both a question decomposition process and a semantic reasoning process with general pseudo code by a zero-shot prompt. HoPC helps to understand semantic logic in the reasoning process and gets a correct answer without requiring any extensions. 

We evaluate our HoPC on GPT-3.5-Turbo, Qwen2.5-7b \cite{Qwen2024Qwen25technicalreport} and Llama3-8b \cite{grattafiori2024Llama3herdmodels}. Our HoPC shows on-par performance with zero-shot CoT on four arithmetic reasoning tasks and one commonsense reasoning task. We also show logical reasoning example. Totally, the contributions of this paper can be summarized as follows:
\begin{itemize}
    \item  We design a formatted instruction chain called Hint Chain of Pseudo Code. In this instruction chain, we first decompose the question into sub-questions and produce general pseudo code on each of them.
    \\
    \item Compare to the previous methods, our HoPC shows a clearer ``mind map'' of LLMs reasoning without any extensions. This helps us to understand the reasoning process better when either correct or wrong.
    \\
    \item Our experiments on four arithmetic reasoning tasks and one commonsense task show that LLMs executing general pseudo code as semantic code interpreters improve performance in tackling reasoning tasks compared with previous methods. 
\end{itemize}

\section{Background}
\subsection{Large language models and prompting}
A language model (LM) is a model that is designed to estimate the probability distribution of text. In recent research, they found that scaling up the model size can help improve the performance (from a few million \cite{merity2016pointer} to hundreds of millions \cite{devlin-etal-2019-bert} to hundreds of billions \cite{brown2020language} parameters). And the training data also becomes more extensive, e.g., webtest corpora \cite{gao2020pile}. These improve the abilities of pre-trained LLMs in many downstream NLP tasks. Unlike the classic paradigm of ``pre-train and fine-tune'', an LLM that scales to 100B+ parameters displays the ability to few-shot learning \cite{brown2020language}. Using in-context learning, we can use prompts to strongly lead the generation to output a desired answer to a specific task. It is called pre-train and prompts \cite{liu2021pretrain}. 

\subsection{Zero-shot prompting}
Based on the drawback of the few-shot CoT that costs time and people to design the prompt, \cite{NEURIPS2022_8bb0d291} proposed a zero-shot CoT prompt. They added \textit{``Let's think step by step''} or a similar text. This work showed that the potential of LLMs is zero-shot. Compared with standard zero-shot and zero-shot CoT, the latter significantly improved with GSM8K from 12.5\%  to 40.5\%. However, \textit{``Let's think step by step''} is a settled prompt that can not be flexible in every reasoning task. This does not inspire the LLM to generate a reasoning process but only a problem decomposition.

To make more accurate calculations on math tasks. PoT \cite{chen2023program} proposed the use of an extended tool to calculate the answer. In their work, they ask the LLM to propose and extend the output Python code to get the math answer. In their zero-shot experiment, PoT performs better than zero-shot CoT on GSM8K (57.0\%), AQuA (43.9\%), SVAMP (70.8\%), TabMWP (66.5\%), MultiArith (92.2\%), and Avg (66.1\%). However, this program-based work does not fit for semantic reasoning tasks because not all problems can be described in code easily. These inspired us to invent a zero-shot prompt that can do problem decomposition and logical reasoning with hints.

\section{Method}

In this section, we indicate Hint of Pseudo Code (HoPC). To increase the reasoning process of zero-shot CoT and the semantic understanding of PoT, we propose Hint of Pseudo Code (HoPC) prompting that encourages LLMs to generate decomposed reasoning paths with pseudo code to represent semantic code reasoning and activate more profound reasoning. In our work, hints as a hint chain and the question are composed as input to LLMs, and the LLMs themselves are required to generate explainable semantic-level step by step reasoning steps along with general pseudo code without any extensions. This employs a chain-like instruction to activate more profound reasoning generation within the LLM. Consequently, the input prompting to the LLM is ``$[X]$. Answer: $hint\_chain$'' where $[X]$ is the question, and $hint\_chain$ is our HoPC prompting.

\begin{figure*}[htbp]
\centerline{\includegraphics[scale=0.38]{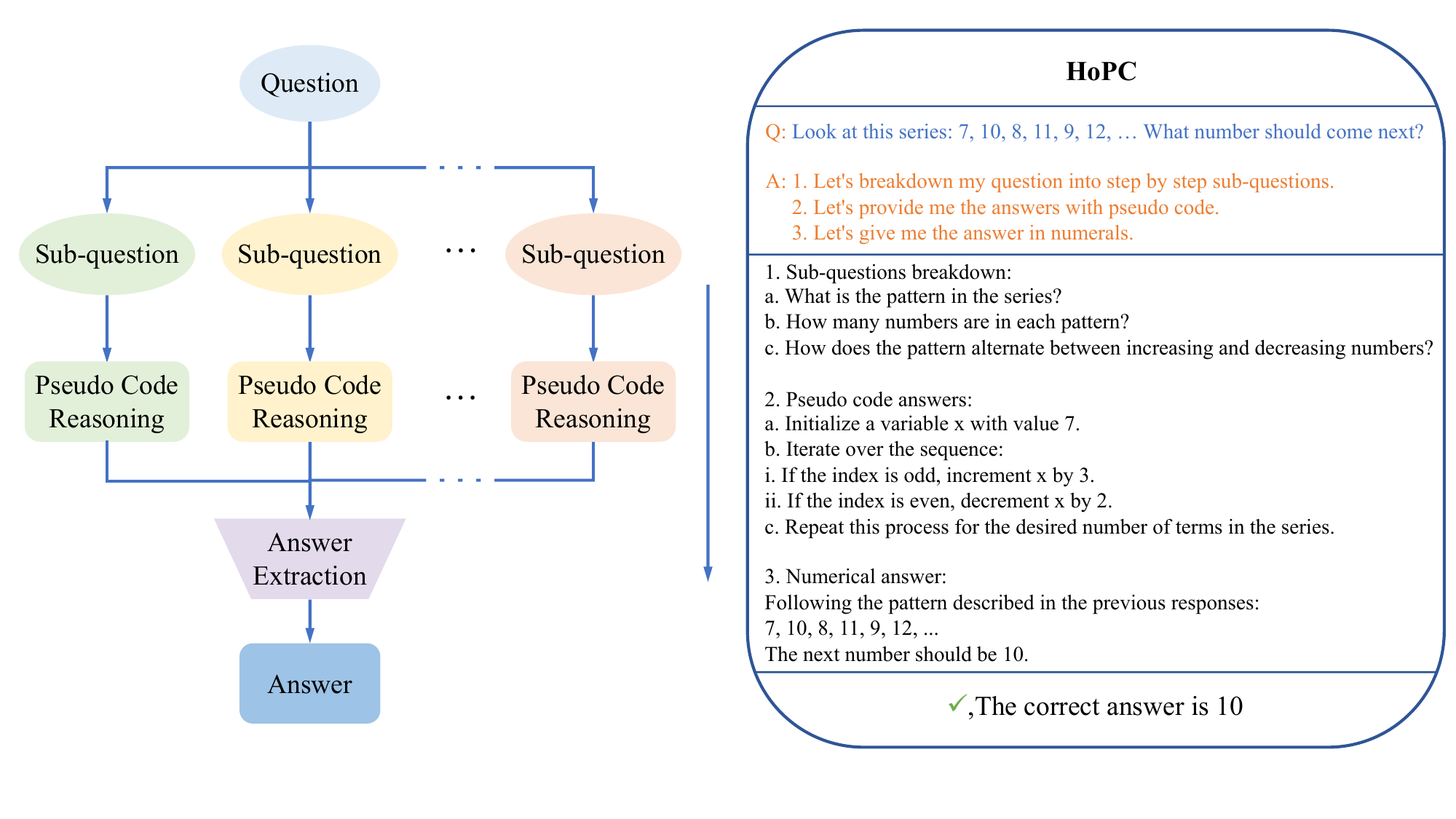}}
\caption{An Example output of HoPC, the orange part is our HoPC prompt and it shows the problem decomposition, the pseudo code reasoning, and the answer extraction. For every questions, we firstly decompose them into sub-questions and do pseudo code reasoning on each sub-questions. Lastly, we combine every pseudo code reasoning and do an answer extraction to achieve the final answer.}
\label{fig3}
\end{figure*}

\subsection{Flexible hint chain}
The hint chain is a chain of instructions that delivers the LLM with instructions step by step. The purpose of the hint chain is to tell the LLM how to generate a semantic code reasoning chain with problem decomposition. There are three parts of such hint chains: (1) Sub-questions: ask the LLM to partition the question into step by step sub-questions; (2) Logical reasoning: tell the LLM to use the step by step sub-questions to generate a pseudo code reasoning process; (3) Answering: ask the LLM to give a customized format of the answer. The result produced by the hint chain follows the order of these three parts and provides a consequential reasoning chain that supports its answer.

\begin{table*}[htbp]
\caption{Example Hint Chain on different tasks. Different hint chains are used to specify the format of the answer. For example, we require numerals answers in arithmetic tasks and Yes/No answers from Yes/No tasks.}
\begin{center}
\begin{tabular}{|c||p{8cm}|}
\hline
\textbf{Question types}&\textbf{Hint Chain}\\ [0.5ex] 
\hline\hline
Arithmetic questions & 1. Let's breakdown my question into step by step sub-questions. 2. Let's provide me the answers with pseudo code. 3. Let's give me the answer in numerals.  \\

\hline

Yes/No questions & 1. Let's breakdown my question into step by step sub-questions. 2. Let's provide me the answers with pseudo code. 3. Let's give me the answer in Yes/No. \\

\hline

General questions & 1. Let's breakdown my question into step by step sub-questions. 2. Let's provide me the answers with pseudo code. 3. Let's give me the answer. \\

\hline

\end{tabular}
\label{tab1}
\end{center}
\end{table*}

\subsection{Explainable question decomposition}
Although the LLM is in the back box representation, our HoPC can produce an understandable and explainable consequent problem-solving procedure. It indeed helps the users verify the answer more intuitively. Question decomposition helps the LLM solve the whole problem and transparently explain the LLMs' mind map. Therefore, a question decomposition in the hint chain can encourage the LLM to explain the solving process.

\subsection{Logical pseudo code reasoning}
An interpretable problem-solving chain can enhance LLMs' ability to solve reasoning problems more accurately. However, although LLM is designed to work with language, it is likely to have semantic ambiguity in semantic-level reasoning. Therefore, pseudo code, regarded as a more accurate logical language in programming design, is used in HoPC. In HoPC, we used general pseudo code. A general pseudo code is a high-level, language-agnostic description of a process or algorithm, designed to illustrate the logical flow of a solution without strict adherence to control structures (e.g., loops, conditionals) or programming syntax. Its primary goal is to communicate ideas clearly and intuitively, using plain language, step by step breakdowns, or conceptual annotations. As a semantic code reasoning, it can produce better reasoning performance on both semantic and arithmetic reasoning problems.

\section{Experimental Setup and Results}

\begin{figure*}[htbp]
    \centerline{\includegraphics[scale=0.4]{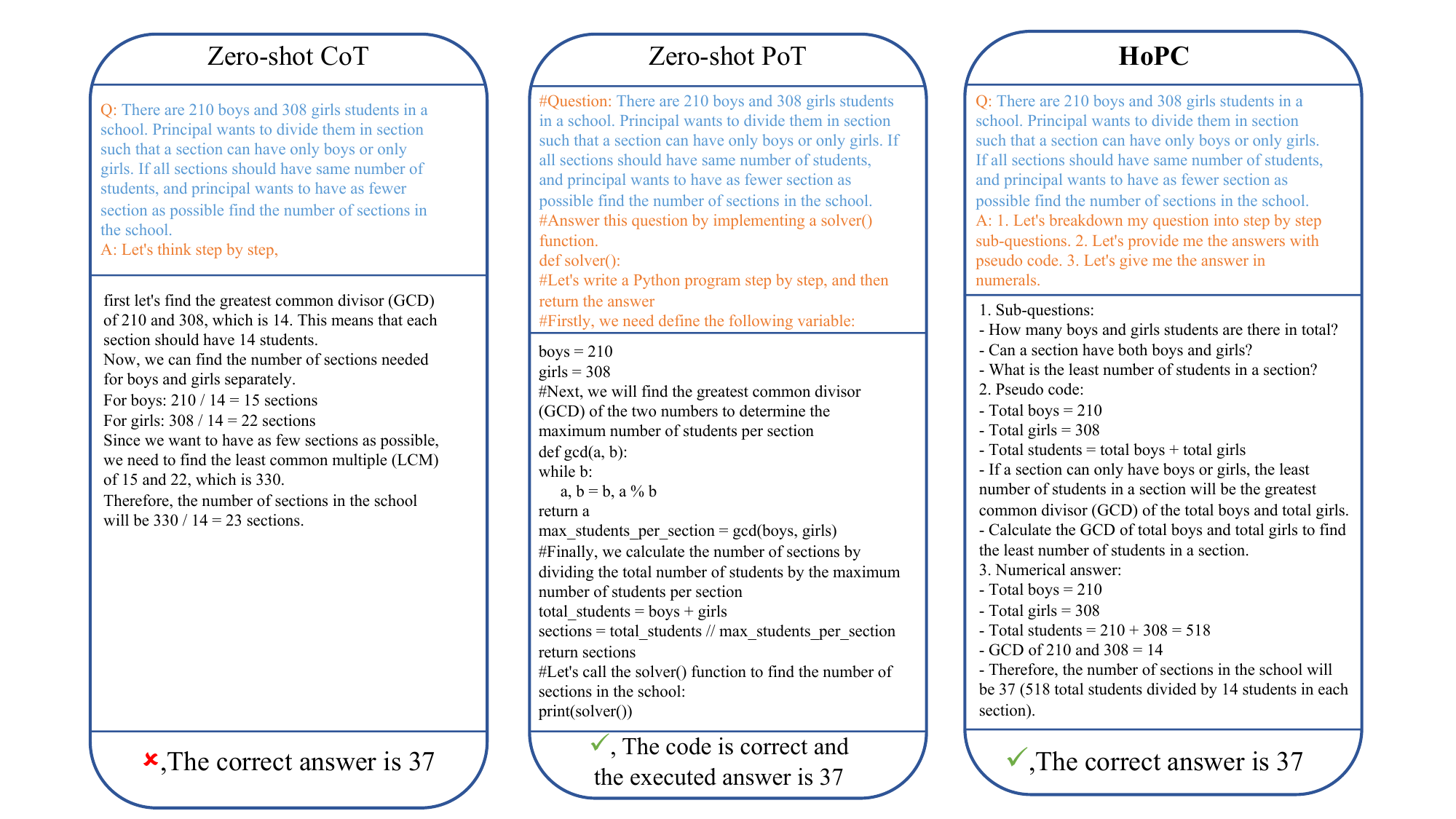}}
    \caption{An arithmetic example of zero-shot CoT, zero-shot PoT and HoPC. These three examples show three different approaches to reach the same arithmetic question. In zero-shot CoT, it goes through the questions briefly and do little calculations to achieve the output which is wrong. In zero-shot PoT, it provides a good written program and a correct answer. However, it is with a long prompt and requires an extra interpreter. In our HoPC, we use a relatively short prompt and generates a long reasoning process with a good structure. Our HoPC with a more precise and structured reasoning process, it gets a correct answer without extra helps. Our HoPC, without an extra program interpreter, but with a more precise and structured reasoning process, gets a same correct answer as PoT does. Note: the orange parts are the prompts.}
    \label{fig5}
\end{figure*}
\begin{figure*}[htbp]
    \centerline{\includegraphics[scale=0.4]{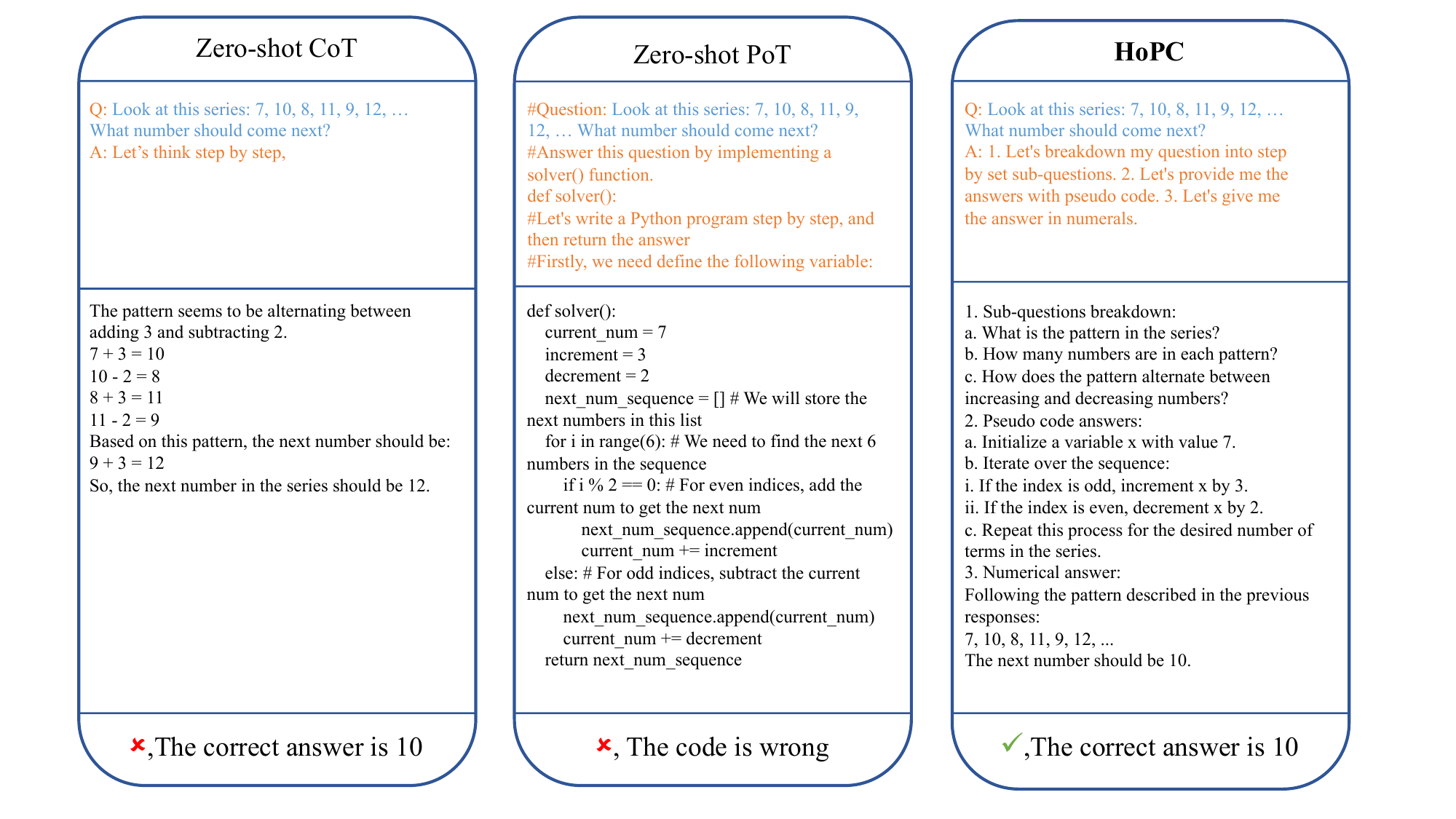}}
    \caption{A logical reasoning example of zero-shot CoT, zero-shot PoT and HoPC. These three examples show three different approaches to reach the same question. In zero-shot CoT, the lacking of structured reasoning process is resulted in a short and wrong output. In zero-shot PoT, it gives a well written python code, however, this code does not provide a final correct answer. In Our HoPC, it generates reasoning processes by decompose the questions and provides pseudo code reasoning on each of sub-questions. Our HoPC with a more precise and structured reasoning process, gets a correct answer. Note: the orange parts are the prompts.}
    \label{fig4}
\end{figure*}
\begin{figure*}[htbp]
    \centerline{\includegraphics[scale=0.3]{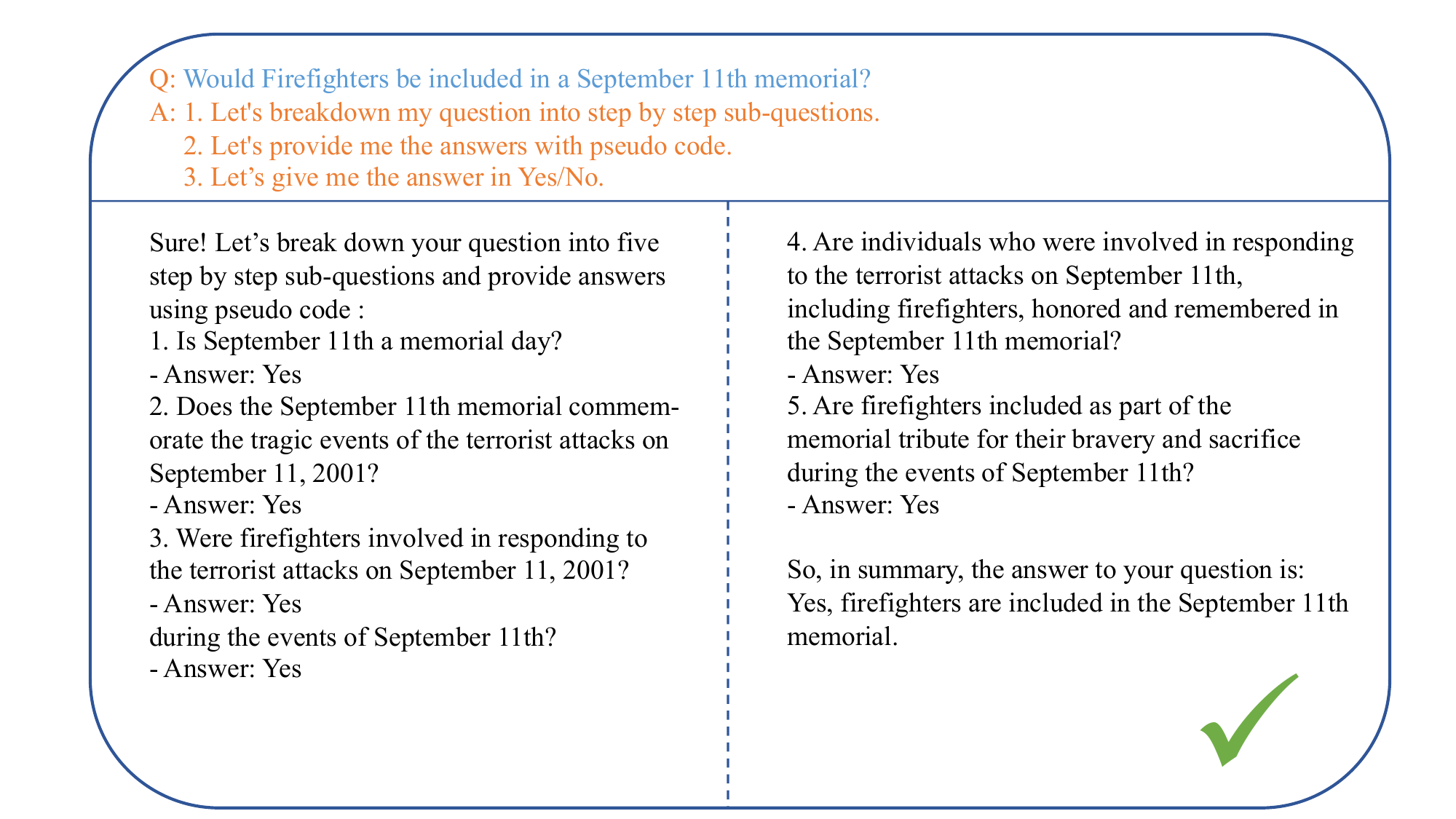}}
    \caption{A commonsense reasoning example of our HoPC. The orange part shows our HoPC prompt. In commonsense reasoning, general pseudo code in here is ``Answer: Yes''. This shows the advantage of general pseudo code that it is not specified in syntax but illustrate the reasoning flow. }
    \label{fig_common}
\end{figure*}

All the experiments done with our HoPC are based on GPT-3.5-Turbo, Qwen2.5-7b \cite{Qwen2024Qwen25technicalreport} and Llama3-8b \cite{grattafiori2024Llama3herdmodels}. Our baselines are adopted from zero-shot CoT \cite{NEURIPS2022_8bb0d291} and PoT \cite{chen2023program}. A contradiction of reasoning is shown in \Cref{fig4} and \Cref{fig5}. We evaluate the HoPC prompting on five datasets for the four main arithmetic reasoning tasks: GSM8K, AQUA, SVAMP, and ADDSUB. In addition, we complete experiments on the big commonsense reasoning benchmark, StrategyQA. GSM8K \cite{cobbe2021training} is a dataset of high-quality linguistically diverse grade school math word problems designed to test the problem-solving abilities of AI systems. AQUA-RAT (Algebra Question Answering) \cite{ling-etal-2017-program} is a dataset designed to evaluate AI systems on solving algebraic word problems, featuring multiple-choice questions that require reasoning and step by step solutions. SVAMP \cite{patel-etal-2021-nlp} is a benchmark dataset designed to test the robustness of AI models on elementary math word problems by introducing simple linguistic and structural variations to existing problems. ADDSUB \cite{hosseini-etal-2014-learning} is a dataset of elementary-level addition and subtraction word problems designed to evaluate AI models' ability to solve basic arithmetic tasks in natural language contexts. StrategyQA \cite{geva-etal-2021-aristotle} is a benchmark dataset designed to evaluate AI models' ability to perform multi-step reasoning and implicit knowledge retrieval to answer complex yes/no questions. Note that all the datasets we utilize in our experiments are publicly released.

\begin{table*}[htbp]
\caption{Experiment on four arithmetic reasoning tasks and one Commonsense reasoning task. Our main experiment of HoPC is done on GPT-3.5-Turbo. The main experiment baseline results of PoT are adopted from \cite{chen2023program}. The * means that this method used special answer trigger depending on answer format, such as ``The answer (arabic numerals) is''. In our Qwen2.5 and Llama3 setups, we used Qwen2.5-7b and Llama3-8b. Our HoPC achieve higher scores than CoT and PoT. It is interesting to find out that, in our Llama3-8b setup, zero-shot CoT harms the performance while HoPC does not. }

\centering
\begin{tabular}{|c|c||c|c|c|c|c|}
\hline
\textbf{Model}&\textbf{Methods}&\textbf{GSM8k}&\textbf{AQUA}&\textbf{SVAMP}&\textbf{ADDSUB}&\textbf{StrategyQA} \\ [0.5ex] 
\hline\hline
GPT & Zero-shot & 21.0 & 45.7 & 66.3 & 76.2 & 52.6 \\

& Zero-shot CoT & 47.0 & \textbf{46.9} & 72.6 & 75.9 & 59.2 \\

& Zero-shot CoT* & 50.3 & 40.5 & 73.5 & 76.5 & 47.5 \\

& Zero-shot PoT & 57.0 & 43.9 & 70.8 & - & - \\

& \textbf{HoPC (Ours)} & \textbf{70.7} & 46.4 & \textbf{76.9} & \textbf{87.3} & \textbf{82.9} \\
\hline

Qwen2.5 & Zero-shot & 1.81 & 30.3 & 14.8 & 13.7 & 59.7 \\

& Zero-shot CoT & 84.8 & 69.7 & 91.7 & 88.6 & \textbf{67.4} \\

& \textbf{HoPC (Ours)} &  \textbf{85.7} & \textbf{70.9} & \textbf{91.8} & \textbf{89.5} & 59.8 \\

\hline
Llama3 & Zero-shot & 1.8 & 30.3 & 3.8 & 4.3 & 61.7 \\

& Zero-shot CoT & 51.8 & 37.4 & 51.8 & 53.6 & 21.0 \\

& \textbf{HoPC (Ours)} &  \textbf{63.5} & \textbf{74.4} & \textbf{75.5} & \textbf{74.9} & \textbf{61.9} \\
\hline

\end{tabular}
\label{tab2}

\end{table*}

\subsection{Arithmetic tasks}
For the arithmetic reasoning task, the following four benchmarks are considered: i.e., 1) GSM8K \cite{cobbe2021training}, 2) AQUA-RAT \cite{ling-etal-2017-program}, 3) SVAMP \cite{patel-etal-2021-nlp}, and 4) AddSub \cite{hosseini-etal-2014-learning}. The first three datasets, especially GSM8K, are more recently published benchmarks and have more challenges because they require multi-step reasoning to solve problems. Also, the AQUA-RAT (AQUA) is a multiple-choice question dataset. We can observe that HoPC performs the best on average among the zero-shot reasoning prompts on various models.

\subsection{Commonsense tasks}
We employ StrategyQA \cite{geva-etal-2021-aristotle} for the commonsense reasoning task, a large dataset that requires the model to conduct implicit multi-hop reasoning to answer questions. It is found out that zero-shot CoT has a lower performance than zero-shot on Llama3-8b, while HoPC can provide a better reasoning performance.

\section{Ablation Study}
We remove the sub-questions and pseudo code parts separately and perform a GSM8K test on GPT-3.5-Turbo. The result shows that HoPC with only sub-questions performs similarly to zero-shot CoT in terms of accuracy and interoperability. Meanwhile, HoPC with only pseudo code produces accuracy similar to that of HoPC but with less interoperability, only outputting pseudo code. This experiment shows that the pseudo code part gives the logical reasoning ability of HoPC, and the sub-questions part gives interoperability.
\begin{figure}[htbp]
\centerline{\includegraphics[scale=0.45]{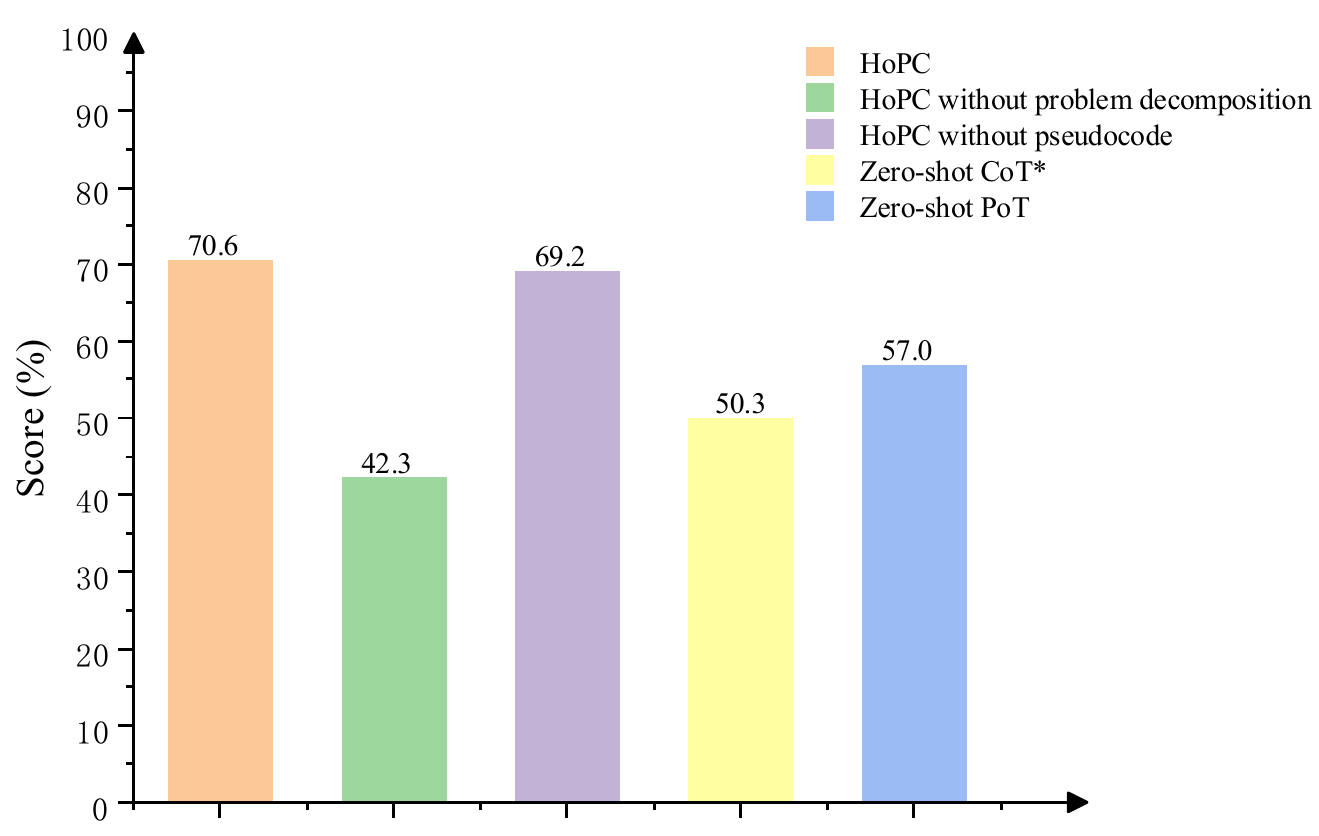}}
\caption{Performance of HoPC with or without problem decomposition and pseudo code. Our work is mainly based on generating step by step pseudo code which means it depends on the problem decomposition ability and pseudo code reasoning ability. In this chart, we can see that step by step question decomposition takes a key role in performance while the pseudo code reasoning also helps in the reasoning process. }
\label{fig7}
\end{figure}

\section{Related Work}
\subsection{Complex reasoning with LLMs}
Reasoning skills are essential for general intelligence systems, and the ability to reason in LLMs has gained significant attention from the research community. Several studies \cite{brown2020language,cobbe2021training} have shown that asking pre-trained models to produce step by step reasoning or fine-tuning can increase their ability on complex reasoning tasks.
GPT-3 \cite{brown2020language} has illustrated its robust few-shot reasoning that a few examples in natural language are given to the model to describe the task. The most classic reasoning task is mathematical reasoning. PoT \cite{chen2023program} has shown great ability on math reasoning tasks with LLMs with the help of Python programs. They aim to generate an executive Python program for the LLM to solve math problems. However, their work primarily focuses on math reasoning tasks. A more general approach would be CoT \cite{wei2023chainofthought}, which works well on mathematical, logical, common sense, and symbolic reasoning tasks with few-shot prompts. 

\subsection{Zero-shot reasoning with LLMs}
It was indicated that LLMs have excellent zero-shot abilities in many system-1 tasks, including reading comprehension, translation, and summarization \cite{Radford}. This ability can also be fine-tuned to get a better performance. However, we focus on system-2 tasks beyond system-1 tasks. The recent work, zero-shot CoT, increases zero-shot performance. Also, PoT in zero-shot format provides good results in math reasoning tasks. 

\subsection{Discussion about existing work}
Recently, there have been many approaches to enhance the reasoning ability of LLMs, including CoT \cite{wei2023chainofthought}, zero-shot CoT \cite{NEURIPS2022_8bb0d291}, Auto-CoT \cite{shin-etal-2020-autoprompt}, PoT \cite{chen2023program}, and decomposed prompting \cite{khot2023decomposed}. They all aim to provide accurate reasoning results. On the other hand, they did not focus on explanations of answers and the LLM's efficiency. So, HoPC is proposed as another approach to prompt engineering.

\section{Conclusion and Limitation}
In our work, we approach reasoning tasks from other perspectives. We care about accuracy and want a more precise explanation of the reasoning process. Therefore, HoPC use a hint chain to guide LLMs to generate pseudo code for sub-questions and execute them step by step.  We have verified that HoPC can work efficiently on arithmetic and common-sense reasoning tasks and provide a clear description intuitively. Although LLMs are black boxes, we tried to explain them with a prompt approach.

This paper investigates how to generate more intuitively explainable and flexible reasoning task prompts. Our experimental results on the various reasoning tasks significantly show the potential of HoPC prompting. 

Our work is based on open-sourced LLMs, pre-trained language models trained from various sources and shown to capture and amplify biases found in the training data. We use prompting to guarantee our reasoning answers, which takes advantage of the patterns learned by language models. However, our zero-shot approach directly probes complex reasoning inside pre-trained LLM, which can also cause bias. Also, our ability to reason is based on the power of the LLM. Therefore, the accuracy depends on LLMs, which may cause fluctuation in different test environments.

\section{Acknowledgment}
This work was supported in part by the National Science Foundation of China (NSFC) under Grant No. 62176134 and by a grant from the Assisted Medical Consultation Project Based on DeepSeek.

\bibliographystyle{splncs04}

\bibliography{main}

\newpage
\appendix
\section{Examples}

\begin{figure*}[htbp]
    \centerline{\includegraphics[scale=0.4]{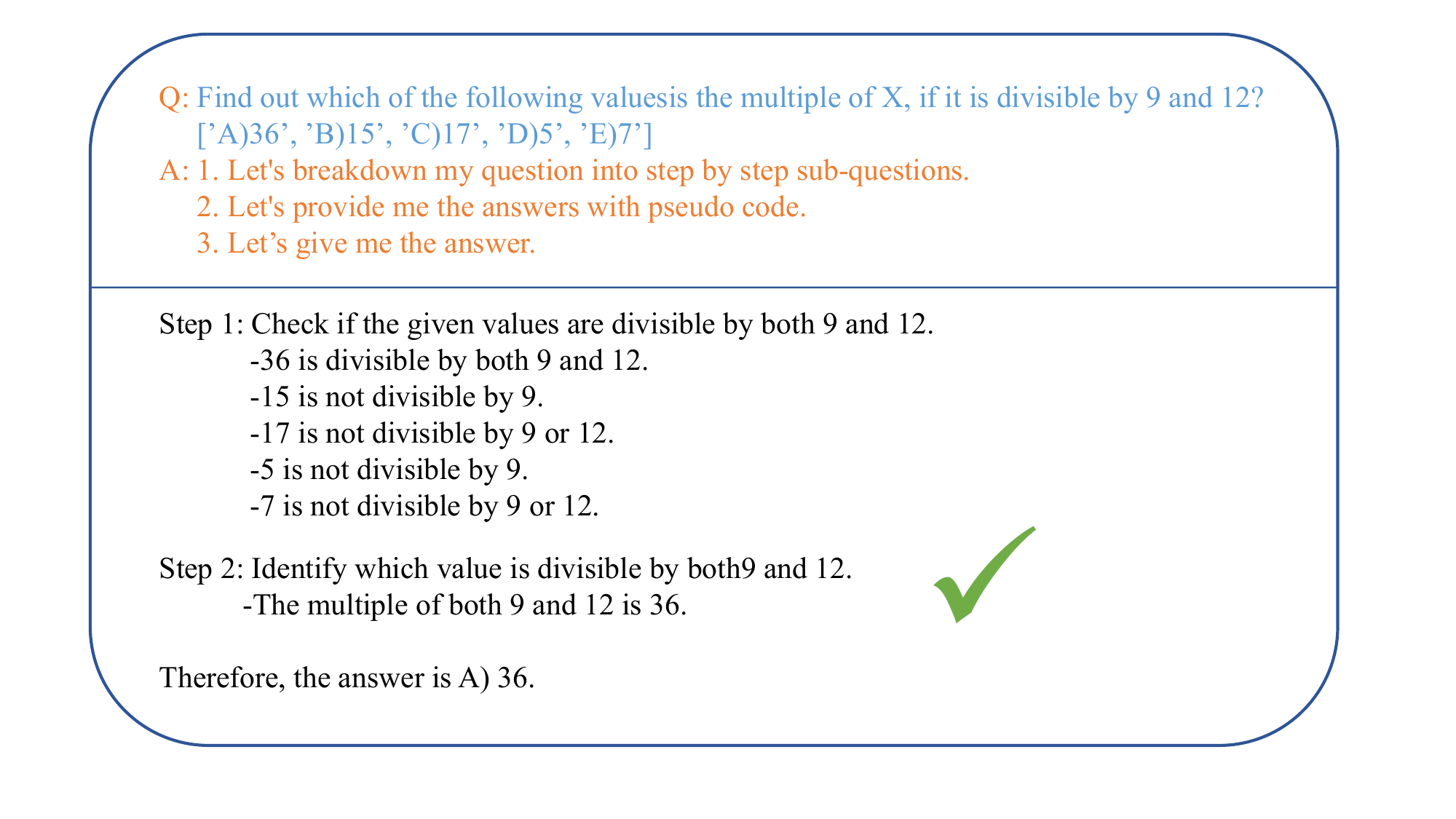}}
    \caption{An AQUA reasoning example of our HoPC. The orange part shows our HoPC prompt. This shows the advantage of general pseudo code that it is not specified in syntax but illustrate the reasoning flow. }
    \label{fig_aqua}
\end{figure*}

\begin{figure*}[htbp]
    \centerline{\includegraphics[scale=0.4]{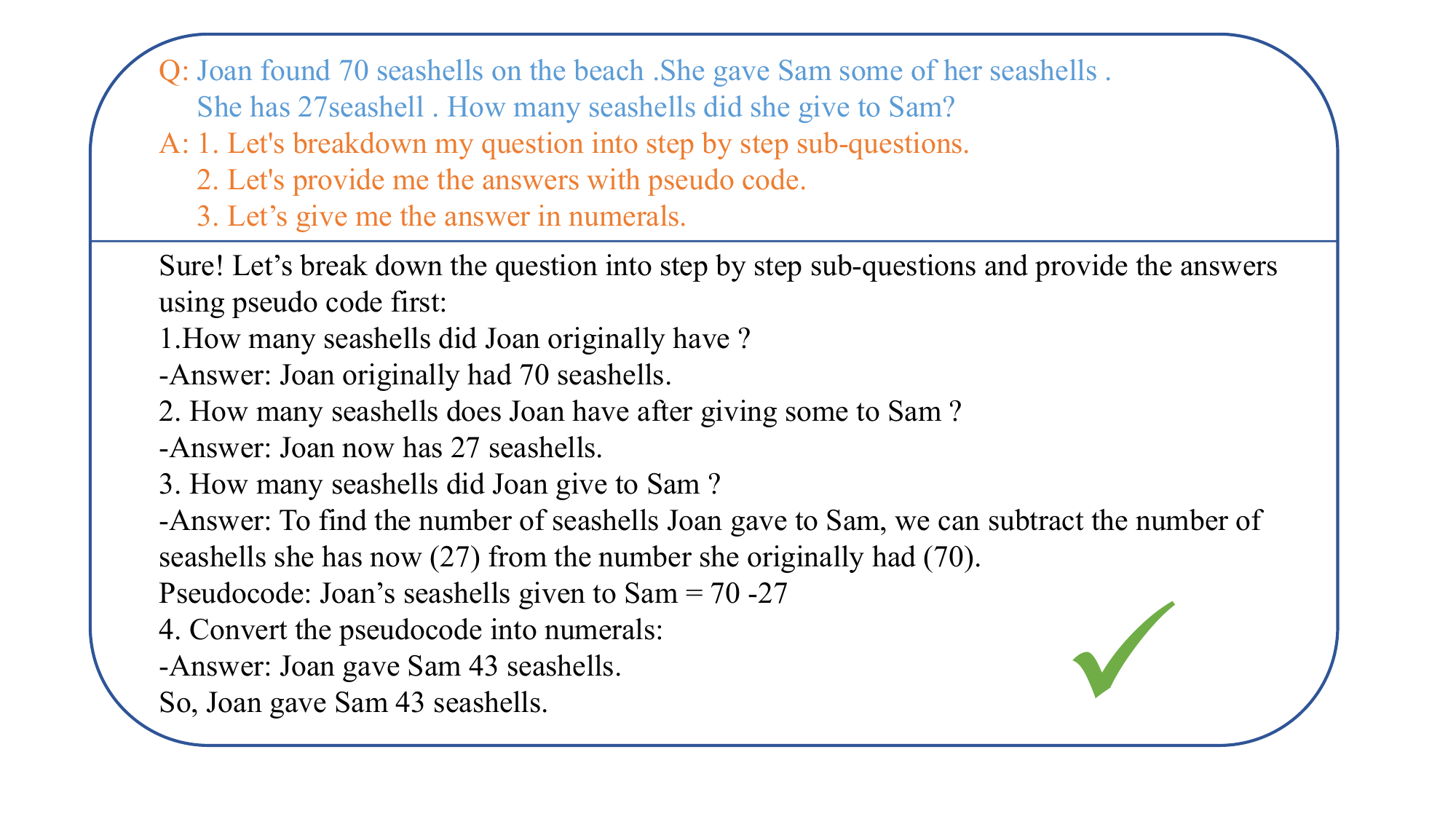}}
    \caption{An ADDSUB reasoning example of our HoPC. The orange part shows our HoPC prompt. This shows the advantage of general pseudo code that it is not specified in syntax but illustrate the reasoning flow. }
    \label{fig_addsub}
\end{figure*}
\begin{figure*}[htbp]
    \centerline{\includegraphics[scale=0.4]{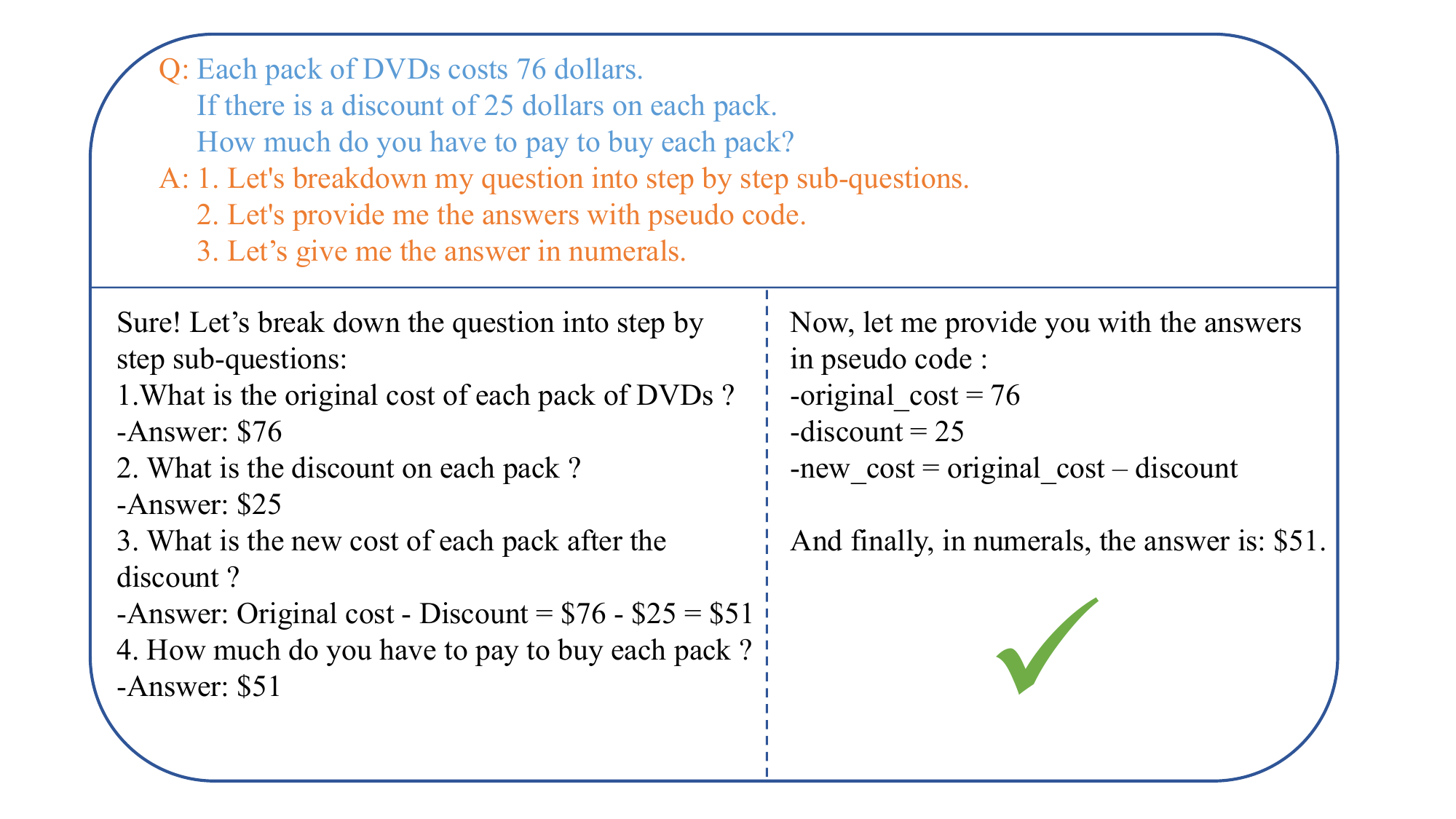}}
    \caption{A SVAMP reasoning example of our HoPC. The orange part shows our HoPC prompt. This shows the advantage of general pseudo code that it is not specified in syntax but illustrate the reasoning flow. }
    \label{fig_svamp}
\end{figure*}


\end{document}